# Review and Comparison of Commonly Used Activation Functions for Deep Neural Networks

Tomasz Szandała[1]

[1]Wroclaw University of Science and Technology
Wroclaw, Poland

The primary neural networks decision-making units are activation functions. Moreover, they evaluate the output of networks neural node; thus, they are essential for the performance of the whole network. Hence, it is critical to choose the most appropriate activation function in neural networks calculation. Acharya et al. (2018) suggest that numerous recipes have been formulated over the years, though some of them are considered deprecated these days since they are unable to operate properly under some conditions. These functions have a variety of characteristics, which are deemed essential to successfully learning. Their monotonicity, individual derivatives, and finite of their range are some of these characteristics (Bach 2017). This research paper will evaluate the commonly used additive functions, such as swish, ReLU, Sigmoid, and so forth. This will be followed by their properties, own cons and pros, and particular formula application recommendations.

deep learning, neural networks, activation function, classification, regression

## Introduction

There exist many applications of deep learning neural network, such as voice analysis, speech or pattern recognition, and object classification. This has been demonstrated in its excellent performance in many fields. Moreover, it comprises a structure with hidden layers; this implies that the layers are more than one [1], [2]. However, from the study of deep learning structure, it can adapt in terms of real-world application with natural scenes due to its salient features. There exist only a few layers in the first deep learning model used for classification tasks; for example, there were only five layers in LeNet5 model[3].

Furthermore, there has been an increased depth due to the need for more advanced network models, applications and available computing power rise[4]. For example, twelve layers in AlexNet[5], nineteen, or sixteen layers in VGGNet depending on variants [6], GoogleNet has twenty-two layers[7], and the largest ResNet architecture has one hundred and fifty-two layers[8]. Finally, Stochastic Depth networks have more than one thousand two hundred layers, which have been trainable proved. Thus, getting deep into neural networks will provide a better understanding of the hidden layers, which will help in boosting their performance and training. A neural cell output in the neural network is calculated by the activation unit. The derivative of the activation function is later used by the backpropagation algorithm. Thus, a differentiable activation function must be picked for analysis. This will enable smooth submission of the function to backpropagation weight updates hence avoiding zigzag formation as in the sigmoid function. Furthermore, Alom et al.[9] suggest that it should be easy to calculate an activation function spare completing power, an essential property in huge neural networks with millions of nodes. Thus, it is evident that the artificial neural network requires activation function as a critical tool in mapping response variable and inputs for non-linear complex and complicated functions. It, therefore, shows that there is an introduction of systems with non-linear properties in different fields[10]. However, the primary role of the activation function is the conversion of an input signal of a node in an A-NN to signal output.

The training on the neural network process becomes difficult and challenging when it has multiple hidden layers. Zigzagging weight, vanishing gradient problem, too complicated formula or saturation problem in the neural network of the activation function are some of these challenges. This leads to a continuous learning process which might take a long time[11], [12]. Byrd et al.[13] discuss a comparison of different activation functions that are made in this research paper both practically and theoretically. These activation functions include softplus, tanh, swish, linear, Maxout, sigmoid, Leaky ReLU, and ReLU. The analysis of each function will contain a definition, a brief description, and its cons and pros. This will enable us to formulate guidelines for choosing the best activation function for every situation.

Thus, this paper is unique since it entails real-world applications of activation functions. Hence, it has a summary of the current trends in the usage and use of these functions against the state of the art research findings in practical deep learning deployments. The complication in this paper will enable suitable decisions about how to choose the appropriate and best activation function and implementing it in any given real-world application. As indicated earlier, it is challenging to manage large test data sets of different activation functions. Thus, Banerjee et al. [14] discuss that tracking experiment process, running different experiments across several machines, and managing training data are some real-world applications of these functions. The analysis of different activation functions with individual real-world applications, followed by a summary is as shown below.

## Base Activation Functions

In the real sense, activation functions can be expressed mathematically in a neural network's node. However, only a few of these functions are commonly used and well known for neural network analysis. Backpropagation algorithm[11], [12] multiplies the derivatives of the activation function. Hence, the picked up activation function has to be differentiable[15]. Furthermore, the function should smoothly submit to backpropagation weight updates to avoid zigzagging, for example, in a sigmoid function[16]. Last, but not least, an activation function should easily calculate the spare computing power, which is an important feature in extremely big neural networks consisting of millions of nodes. Below is the analysis of some functions with individual pros and cons, including real-world applications, which will lead to a critical comparison between them.

## Step Function

The first question that the classification of the activation function should answer is whether or not the activation of the neuron should take place. An individual can only activate the neuron in a situation where the input value is higher than a given threshold value or leave it deactivated when the condition is not met. The figure below demonstrates how a step function can either be activated or deactivated.

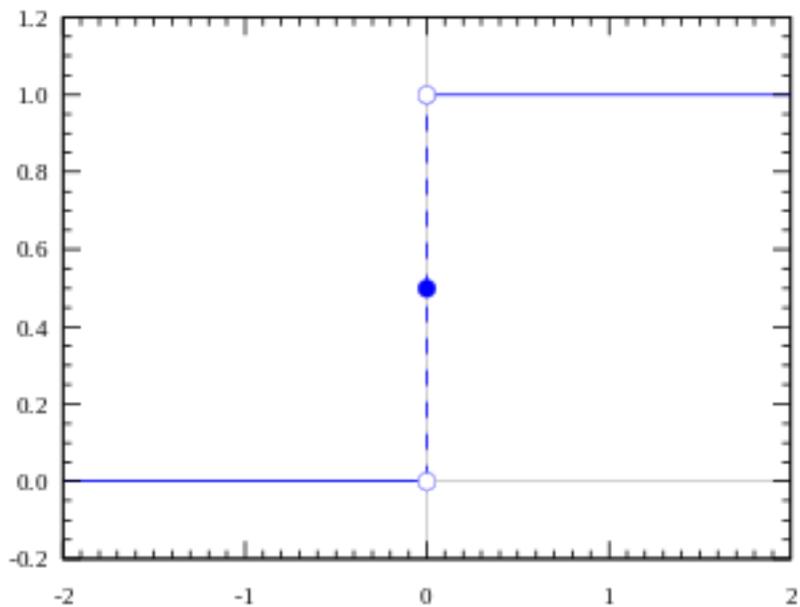

*Fig. 1. Step function*

$$f(x) = \{0,\ for\ x \leq T\ \ 1,\ for\ x > T \quad (1)$$

The above figure follows a function $F(x) = 1$, with conditions: for $x$ and for $x > T$. Thus, neural functions require logical functions to be implemented in them for effective functioning. Moreover, primitive neural networks commonly implement step function, and in this case, they must be with hidden layers, preferably a single layer. He et al. [17] suggest that this is done because the learning value of the derivative is not represented and also it will not have any effect in the future. Hence, the classification of this type of network is under linear separable problems, such as XOR gate, AND gate, and so forth. Therefore, a single linear line can be used to separate all classes 0 and 1 of neural networks. Thus, this function is not flexible hence has numerous challenges during training for network model analysis. Hence, it can only be used as an activation function of a single output network for binary classifications. These functions can only be represented by their derivatives, as shown in the graph below.

## Linear Activation Function

The zero gradients are the major problem in step function; thus, it does not update gradient during the backpropagation as a result of gradient/slope descent being unable to be progressive. Therefore, we can try applying a linear function instead of the step function. In consideration to the simplest example of a linear equation, the equation output can be equivalent to the input, but in more cases, the value of "a" varies with 1, where "a" is the proportional input activation. The equation below demonstrates a simple linear equation with the required variables [Eq. 2.].

$$f(x) = a * x, where\ a \in R \quad (2)$$

Many and various neurons can be activated at the same time using linear activation. In this case, if we decide to choose multiple classes, our interest should be on the one with the maximum value. However, there still exists an activation problem in this situation too. For instance, consider the derivative of the function as shown in equation two below [Eq. 3.].

$$f'(x) = a \quad (3)$$

Also, it appears that issues with gradient descent for training occurs in this function too. Thus, it has a constant derivative function in all situations. This implies that there is no relationship between its x and the gradient. Hence, the constant gradient makes the descent a constant too. Hara, Kataoka, and Satoh [18] suggest that in the case where we have a prediction error, changes are made by the backpropagation, which is constant and is independent of the respective weights used for analysis. Furthermore, this function is of no use in a multi-layer network for deep learning. Thus, a linear function activates each layer. The activation interns are taken to the next level as an input. The weighted sum is calculated by the second layer in the input. This is followed by a firing based on another linear function. In the case where we have a linear function, the last activation function of the final layer remains linear of the input of the first layer, no matter how many layers are added. Therefore, the multiple layers used in activation can be replaced by a single layer, which is a combination of a linear function and will remain linear. However, linear activation even makes sense in the regression prediction of the output layer of networks. Thus, the linear function and its derivatives are expressed on the graph, as shown below.

However, there is still one place where linear activation makes sense: the output layer of networks used for predicting regression[19].

## "S"-shaped Activation Functions

These types of activation functions used to be very common among artificial neural networks. The output of these activation functions is not linear. Thus, the range of the function is 0-1 forming an S shape. These activations functions are discussed below.

## Sigmoid Function

A mathematical function with the features of a sigmoid curve is referred to as a sigmoid function. The combinations of the sigmoid function are not linear since this type of activation function is nonlinear. The figure below represents a demonstration of this function.

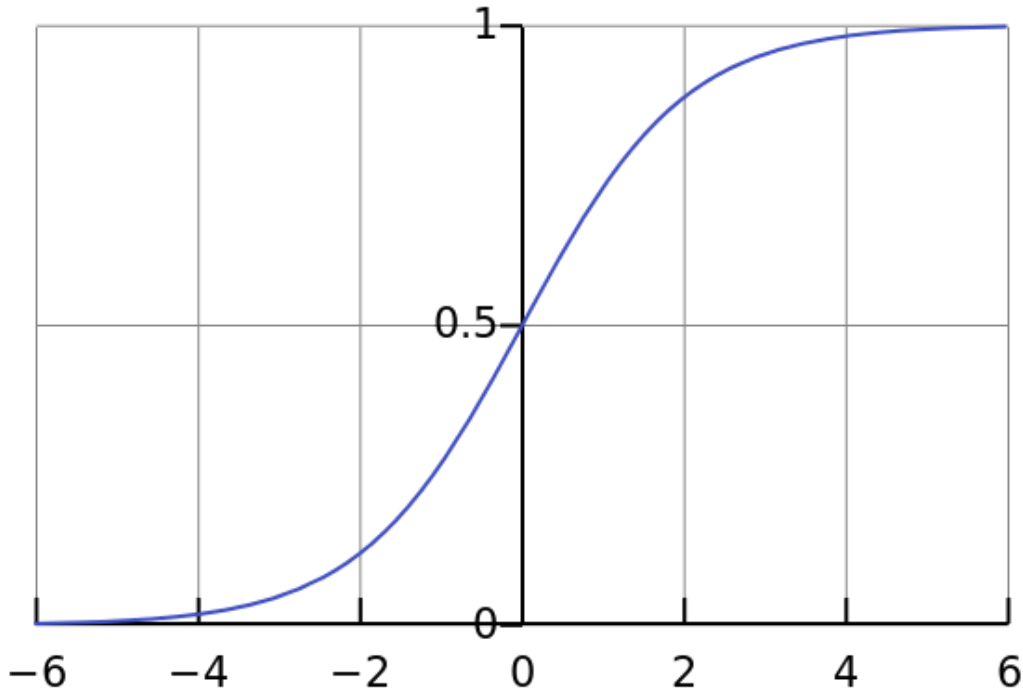

*Fig. 2. The demonstration of a Sigmoid Activation Function[2]*

Thus, it makes sense to stack layers. This applies to non-binary activation as well. It also has a smooth gradient value. Hence, this makes it suitable for shallow networks like functions simulation. The equation [Eq. 4.] below shows how sigmoid functions can be used in deep learning.

$$\sigma(x) = \frac{1}{1+e^{-x}} \quad (4)$$

From the above equation, it can be observed that the Y values are very steep when the X values are between -2 to +2. Thus, there will be a significant change in Y for any slight change in X within this range [2]. This makes the function to have a tendency of bringing the Y values to either end of the curve; thus, enhancing the classifier by making clear prediction distinctions. This makes it even more suitable for shallow networks like logic functions simulation[20].

$$\frac{d}{dx}\sigma(x) = \sigma(x)(1-\sigma(x)) \quad (5)$$

Moreover, another advantage of this function is that its range is $(0, 1)$, unlike the linear function whose range is $(-\infty, \infty)$. Therefore, the range of activation is bounded, and it is easy to prevent blowing up during the activations. Hence, this is the most widely used activation function, and it has been used for a long time. But research shows that sigmoid function is not perfect since the values of Y tend to respond slightly to the changes in X values towards either end of the function. This implies that there will be a small gradient at this point. Thus, this leads to a vanishing gradient problem, which is realized towards the near-horizontal part of the activation functions of the curve on either side. Hence, the gradient vanishes or reduces in size and cannot lead to any significant change since its value is minimal. In this case, the network can either learn drastically slowly or refuse to learn further; this will depend on how it is used until the gradient comes closer to the value limits of the floating-point.

Another problem with the sigmoid function is its non-zero centrality[21]; in other words, it always gives a positive result. Consider a sigmoid unit $y$ with inputs $x_1$ and $x_2$, both from the sigmoid neuron, weighted $w_1$ and $w_2$ respectively. From the definition of a sigmoid function, it is clear that $x_1$ and $x_2$ are always positive, same as y. This generates the error signal which comes into y is during backpropagation, the error gradient will be either positive for w1 and positive w2 or both negative for $w_1$ and $w_2$. Liu et al. [22] demonstrate that when the optimal weight vector calls for an increase in $w_1$ and a decrease in $w_2$, the backpropagation procedure cannot improve both weights simultaneously in a single step, because it must either increase both or reduce both. Thus, the entire training procedure takes more steps to converge than it might need with a better activation function. The sigmoid activation function and its derivatives can also be represented in a graph, as shown below.

## Hyperbolic Tangent Activation Function

The stuck of the neural network on the edge value may occur if we use only the sigmoid activation function. Thus, we need to apply the hyperbolic function as an alternative, which is also known as the tanh function. The figure below[Fig. 3.] shows a comparison between the sigmoid and tanh function.

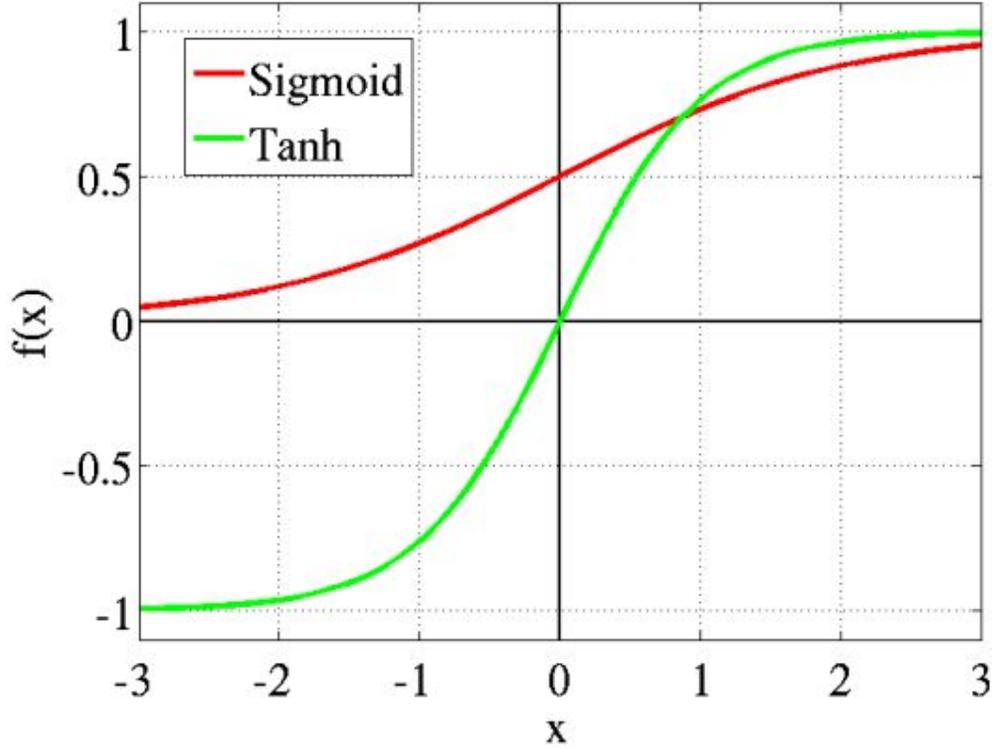

*Fig. 3. Sigmoid and Tanh Functions Comparison Diagram*

Mollahosseini, Chan, and Mahoor[23] suggest that the hyperbolic function is sigmoidal, just like in the case of sigmoid functions. However, for tanh the output values range between −1 and 1, which is just the sigmoid function curve extended. Hence, negative inputs of the hyperbolic functions will be mapped to a negative output as well as the input values that are nearing zero will also be mapped to output values nearing zero. Therefore, the network is not stuck due to the above features during training.

$$f(x) = tanh(x) = \frac{e^x - e^{-x}}{e^x + e^{-x}} \quad (6)$$

Another reason why tanh is preferred to sigmoid is that the derivatives of the tanh are noticeably larger than the derivatives of the sigmoid near zero. However, when it comes to big data, an individual usually struggles to quickly find the local or global minimum, which is a beneficial feature of tanh derivative. In other words, an individual is supposed to minimize the cost function faster if tanh is used as an activation function. Last, but not least, tanh diminishes the zigzagging problem of the only positive sigmoid[16].

$$\frac{d}{dx}tanh(x) = 1 - tanh^2(x) \quad (7)$$

However, a commonly used case is binary classification. Thus, using either sigmoid or tanh activation in the final layer produces a quantity that can be scaled from 0 to 1. Hence, it is suitable to use a function that requires probabilities as inputs, such as cross-entropy. Therefore, tanh, just like sigmoid struggles with the vanishing gradient problem.

## Softsign Activation Function

It is worth mentioning a few attempts to upgrade "s"-shaped functions[24], [25]. Softsign is one of the most important functions, which is smoother than tanh activation function.

$$f(x) = \frac{x}{1+|x|} \quad (8)$$

Moreover, this function grows poly-nominally rather than exponentially[Eq. 9.]. This gentler non-linearity results in better and faster learning due to lack of struggling with vanishing gradient. Thus, researchers have found that Softsign has prevented neurons from being saturated, resulting in more effective learning. However, it is more expensive to compute than tanh; in other words, it has more complex derivatives. Additionally, its gradient sometimes yields extremely low/high values, such that we can consider it as a sigmoid on steroids[26]–[28].

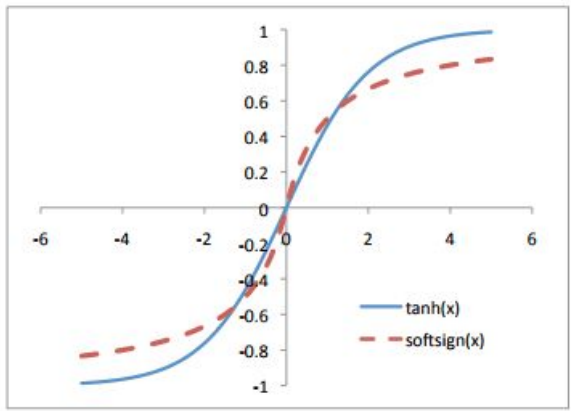

*Fig. 4. Softsign vs tanh function*

$$f(x) = \frac{1}{(1+|x|)^2} \quad (9)$$

## Vanishing Gradient Problem

Just like all "s"-shaped functions, vanishing gradient function has a burden of vanishing gradient[29]. For instance, a sample is taken from MNIST dataset and is classified using 1, 2, 3, and 4 hidden layers each with 30 nodes in it. The sigmoid function is used as the vanishing gradient problem since it is the best visible with this function. Thus, after each training phase, the network is validated against a validating set. The table below [Tab. 1.] shows the accuracy per amount of the hidden layer quantity.

*Table 1. Accuracy per amount of hidden layer quantity*

| Number of hidden layers | Accuracy for Sigmoid |
|---|---|

| 1 hidden layer | 96.81% |
|---|---|
| 2 hidden layers | 97.18% |
| 3 hidden layers | 97.29% |
| 4 hidden layers | 97.10% |

Firstly, the accuracy grows a little for 2 and 3 layers, but at four hidden layers it drops down, close to an extended network. Job classification is performed better due to an extra hidden layer, which enables the network to more complex classification functions[8]. The network is analyzed to determine the insight of what might be wrong with the function. The error or the final layer is propagated back to the previous layer. Each time propagation occurs, a gradient of the gradient is passed, and so a smaller value is also being passed. For this to be proven, there must be defined a common technique of comparing the speed of learning in the hidden layers.

To prove this, we need to have a global way of comparing the speed of learning in each subsequent hidden layer. To do this, we denote the gradient as a derivative of the cost function $\delta_j^l = \partial C / \partial b_j^l$, i.e., the gradient value for the $j$-th neuron in the $l$-th layer. We consider the gradient $\delta^1$ as a vector whose inputs determine how quickly the first hidden layer trains and $\delta^2$ as a vector whose entries determine how quickly the second hidden layer learns. Now, the lengths of these vectors can be utilized as global measures of the speed at which the layers learn. For instance, the length $\|\delta^1\|$ measures the time at which the first hidden layer learns, while the length $\|\delta^2\|$ measures the velocity at which the second hidden layer learns. With this definition, we can compute $\|\delta^1\| = 0.07$ and $\|\delta^1\| = 0.31$. This confirms our previous assumption: the neurons in the subsequent hidden layer really learn much faster than the neurons in the prime hidden layer. Finally, for all 4 hidden layers the respective speeds of training are 0.005, 0.029, 0.091 and 0.345. The pattern is evident: earlier layers learn much slower than later layers. This is even better visible if we analyze the learning rate throughout the entire learning process [Fig. 5.].

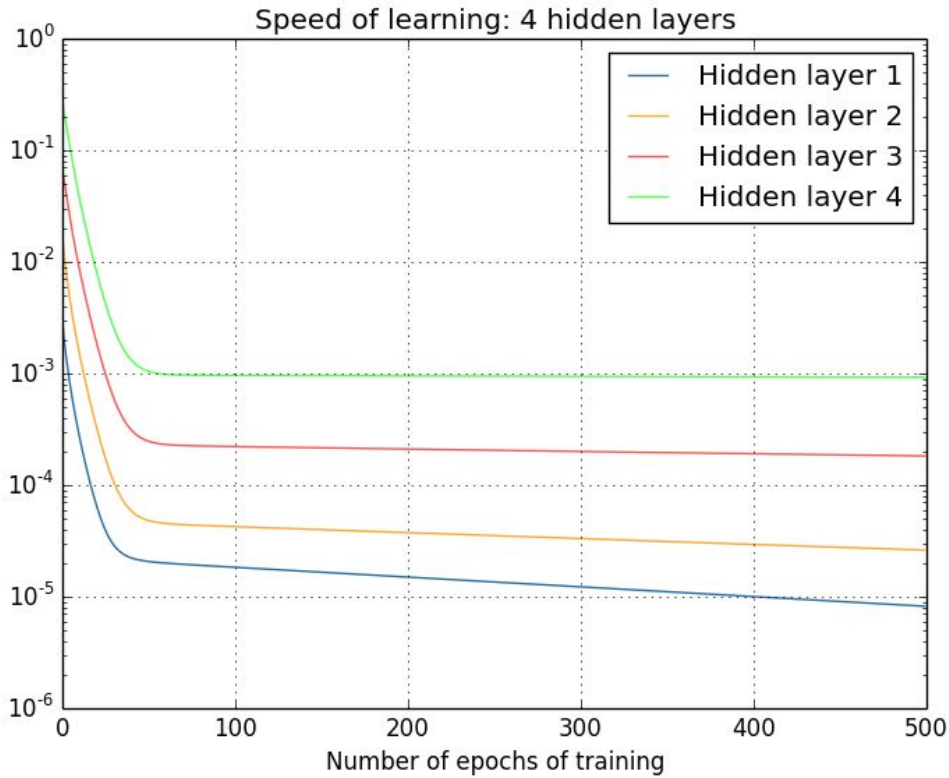

*Fig. 5. Learning speed ( $\|\delta^l\|$ ) for each layer during the learning process*

To sum up, deep neural networks suffer from a problem known as vanishing gradient, which can be avoided by using functions with static or wider-range derivatives.

## Rectified Linear Unit Activation Functions

This type of activation function is responsible for transforming the weighted input that is summed up from the node to the strict output or proportional sum. These functions are piecewise linear functions that usually output the positive input directly; otherwise it the output is zero. Types of rectified linear unit activation functions are discussed below.

## Basic Rectified Linear Unit (ReLU)

The rectifier is an activation that assigns zero to values and value itself is above zero [Eq. 10.]. This is also known as a ramp function and is analogous to half-wave rectification in electrical engineering [Fig. 8.]. However, the activation function was first introduced in a dynamical network with strong biological motivations and mathematical justifications[30], [31].

$$f(x) = x^+ = max(0, x) \quad (10)$$

Moreover, it was proven for the first time in 2011 as enabling the more efficient training of deeper networks, compared to the commonly-used activation functions before 2011, for instance, the sigmoid or the hyperbolic tangent. Ping et al. [32] suggest that the rectifier was the most popular activation function for the deep neural network as per research in 2018. Finally, calculating the function result and gradient is an easy task because the forward and back propagation steps quickly. The results from the research show that ReLU is six times faster than other well-known activation functions[33]. It is noted that for simple regression problems with neural networks, tanh can be superior to ReLU[34]. However, any function approximated with ReLU activation function is always piecewise linear.

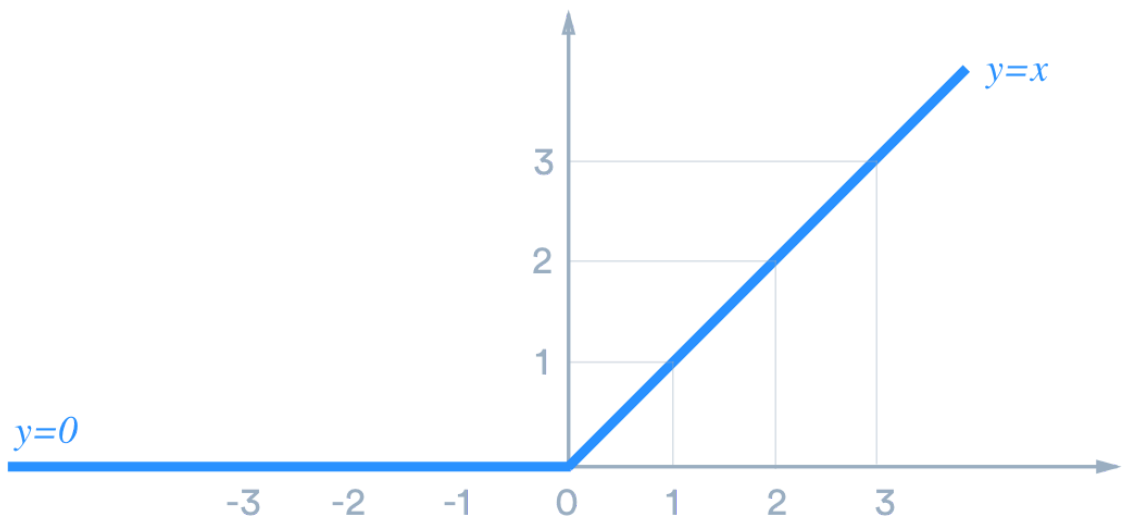

*Fig. 6. Rectifier Linear Unit function*

Moreover, it takes a lot of piecewise linear functions to fit a smooth function like sine. Meanwhile, tanh is very smooth and does not take as many tanh primitives to build something that closely resembles a sine wave. Thus, for classification problems, especially using multiple convolution layers, ReLU and its minor variants are hard to beat. The graph below shows the ReLU function and its derivative[Fig. 7.].

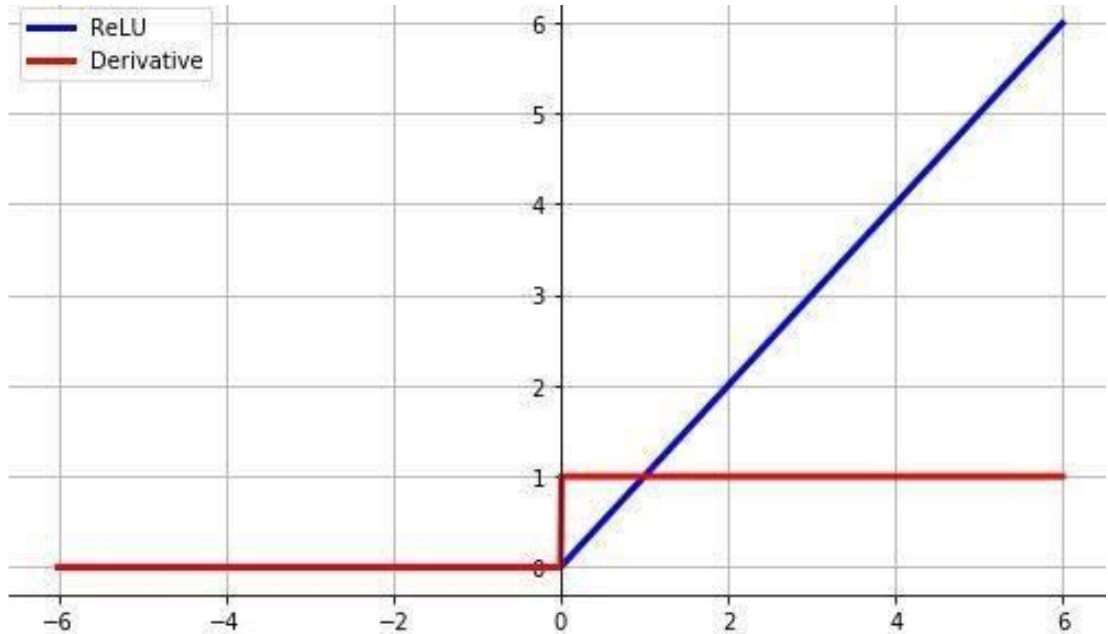

*Fig. 7. ReLU Function and its Derivative*

## Dying ReLU Problem

Unfortunately, ReLU has a significant problem known as "dying ReLU." However, this occurs under certain conditions[35]. Suppose there is a neutral network with weights distribution in the form of a low-variance Gaussian centered at $+0.1$. Thus, by this condition, most inputs are positive and cause ReLU node to fire. Then, suppose that during a particular backpropagation, there is a large magnitude of gradient passing back to the node. Since the gate is open, it will move this large gradient backward to its inputs. This causes a relatively significant change in the function that computes input[36]. Implying that the distribution has moved representing a low variance Gaussian centered at $-0.1$. All the inputs are negative, making the neuron inactive, omitting the weight updates during backpropagation[Eq. 11.]. A relatively small change in "R"s input distribution leads to a qualitative difference in the nodes' behavior. However, the problem is that a close ReLU cannot update its input parameters. One could also consider trying to "recover" dead ReLUs by detecting them and randomly re-initializing their input parameters. However, this would slow down learning but might encourage more efficient use of metrics. Therefore, to bypass this problem, without an additional mechanism, a modification has been proposed, called Leaky ReLU.

$$f'(x) = \{0, \text{ for } x \leq 0 \ \ 1, \text{ for } x > 0 \quad (11)$$

## Dying ReLU - example

Dying ReLU does not always have to appear,, but let's take an example where it may occur. One of the simplest networks is the XOR gate network. However, it is evident that the dying ReLu problem has indeed occurred in the simulations shown below [Fig. 8.].

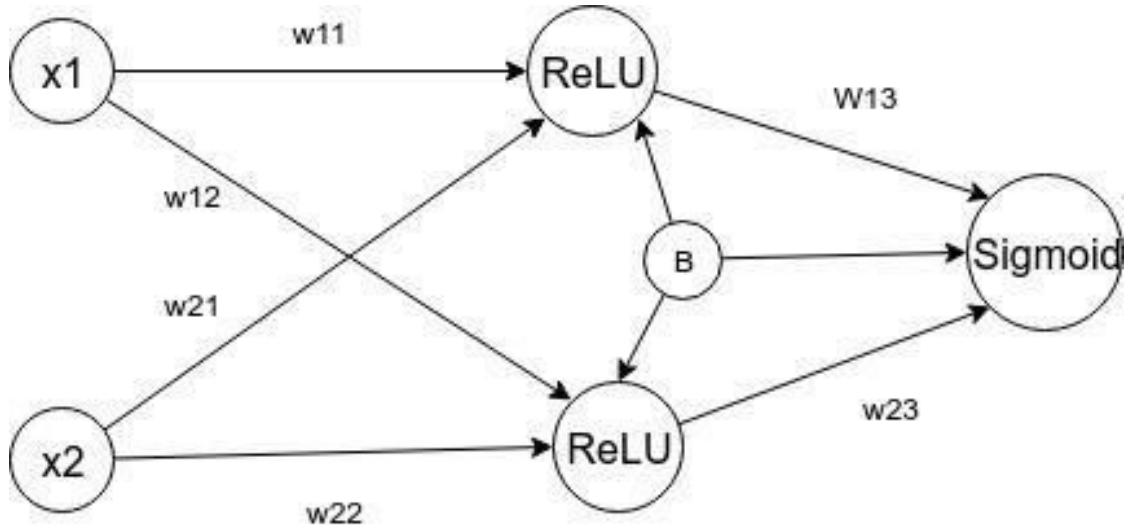

*Fig. 8. Sample network fulfilling XOR gate behaviour*

Because we wanted the output to be either 0 or 1, for output function, we chose sigmoid and ReLUs that were put in the hidden layer nodes. Let's set initial weights as follows[Tab. 2.].

*Table 2. Potential random assigned weights values in the sample XOR network*

| w11 | w12 | b1 |
|---|---|---|
| 0.55 | 0.55 | -1 |
| **w21** | **w22** | **b2** |
| 1 | 1 | -1 |
| **w31** | **w32** | **b3** |
| 1 | 1 | 0.5 |

Weights are chosen randomly for each network, but in this particular example, we adjusted them to invoke 'dying ReLU.' Of course, they are real, potential values that can be assigned. In the table below [Tab. 3.] it can be concluded that the results are not perfect. Therefore, we need to adjust them using backpropagation.

*Table 3. The output of the pre-trained XOR network*

| Input 1 | Input 2 | Expected (*exp*) | Real output(*real*) |
|---|---|---|---|

| 1 | 1 | 0 | 0.83 |
| 1 | 0 | 1 | 0.62 |
| 0 | 1 | 1 | 0.62 |
| 0 | 0 | 0 | 0.62 |

Where in the formula 12.

$$w'_{31} = w_{31} + \eta * f'_3(x) * (exp - real) * f_1(x) \quad (12)$$

we have measured the weight between the first neuron and final neuron, updated by adding an adequate modifier. If we continue to measure all weights as shown in the following formula[Eq. 13.], we receive the following:

$$w'_{11} = w_{11} + \eta * f'_3(x) * (exp - real) * f'_1(x) * w_{13} * input1 \quad (13)$$

Table 4. Post training weights value in sample XOR network

| **w11** | **w12** | **b1** |
|---|---|---|
| 0.49 | 0.49 | -1.05 |
| **w21** | **w22** | **b2** |
| 0.94 | 0.94 | -1.05 |
| **w31** | **w32** | **b3** |
| 0.99 | 0.94 | 0.44 |

Now, we can clearly see that the problem of "dying ReLU" affects neuron one[Tab. 4.]. Therefore, this neuron will never evolve, and we cannot achieve XOR gate behavior. While values for this example chosen, are not impossible, nor disturbing one's prohibition.

## Leaky Rectified Linear Unit (ReLU)

To keep away the dead nodes while using ReLU neurons, we need to reduce the possibility of returning 0 value as output. To achieve this, a small multiplier $\alpha$, e.g., $0.01$, can be introduced for the values below $0$ [37]. This factor causes the inclination of the negative part of a function, therefore preventing proclivity for death[Fig. 9.].

$$f(x) = \{\alpha * x, \text{ for } x \leq 0 \ x, \text{ for } x > 0 \quad (14)$$

Comparing Leaky (also known as Parametric) ReLUs to standard ReLU layer gives no activation value for negative values. Thus, one can take advantage to save some energy in a hardware implementation, like the minimized Internet of Things devices. Still, it does not make a difference when the network is run on a GPU. The graph below pictures a leaky ReLU function and its derivative.

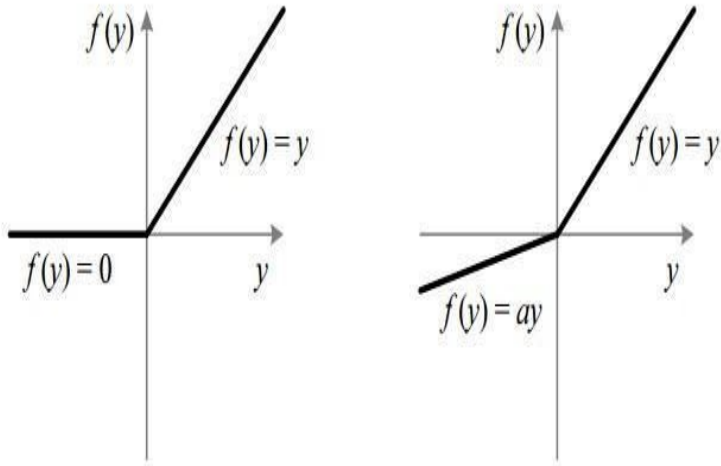

*Fig. 9. ReLU compared to Leaky ReLU function*

## Maxout

For leaky ReLU, it might be hard to choose the parameter for distorting the negative part arbitrarily. Because of that, a more generalized approach has been proposed by the researchers. The Maxout neuron computes the function for the maximum out of two independent weight sets[38].

$$f(x) = max(\vec{w}_1 * \vec{x} + b_1, \vec{w}_2 * \vec{x} + b_2) \qquad (15)$$

However, for every single neuron, Maxout doubles the number of parameters, unlike the ReLU neurons. Moreover, all the benefits of ReLU units, such as no saturation and the linear regime of operations, are embraced by the Maxout neuron. While it can provide the best results among all ReLU-like functions, it comprises the highest cost of training; thus, should be used as the last resort[39]. This 'last resort' solution can be recognized by the phone speech where the Maxout function is successfully applied[40]. The graph[Fig. 10.] below shows the Maxout when $n = 2$.

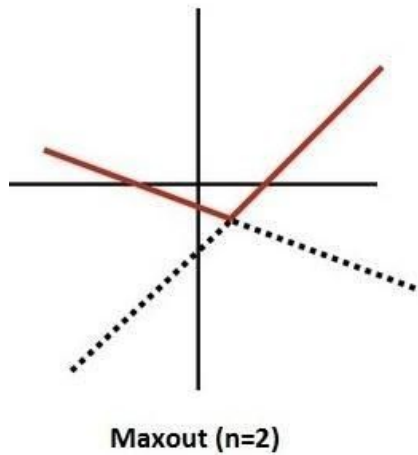

*Fig. 10. The maxout Activation Function n=2*

## Softplus Activation Function

Another alternative to dead ReLU is Softplus. It was first described as a better alternative to *sigmoid* or *tanh*. The output produced by the softplus function ranges from 0 to ∫, whereas the *tanh* and *sigmoid* produces an output with lower and upper limits. The softplus equation is shown below[Eq. 16.].

$$f(x) = ln(1 + e^x) \qquad (16)$$

One of the biggest pros for Softplus is its smooth derivative used in backpropagation. The derivative of Softplus function is equal to sigmoid function. Moreover, the local minima of greater or equal quality, are attained by the network trained with the rectifier activation function despite the hard threshold at zero[41]. However, the derivative of the softplus is equal to a sigmoid function. The graph below shows a softplus function.

## Swish Activation Function

According to research by Google brain team[42], the swish activation function is an alternative to ReLU. Though, for both backpropagation and feed forwarding, the cost is much higher for the computation of the function[43]. Although ReLU has many limitations, it plays a crucial role in deep learning studies. However, research indicates that ReLU is over-performed by the new activation function for the deep network. Swish activation function is represented by the equation below[Eq. 17.].

$$f(x) = x * \sigma(x) = \frac{x}{1+e^{-x}} \qquad (17)$$

Moreover, some authors proposed[43] a non-zero β parameter for enabling a basic swish, the equation for this modification is as shown below[Eq. 18.].

$$f(x) = \beta x * \sigma(\beta x) = \frac{\beta x}{1+e^{-\beta *x}} \quad (18)$$

The figure below shows E-swish producing different graphs using different values of beta[Fig. 11.].

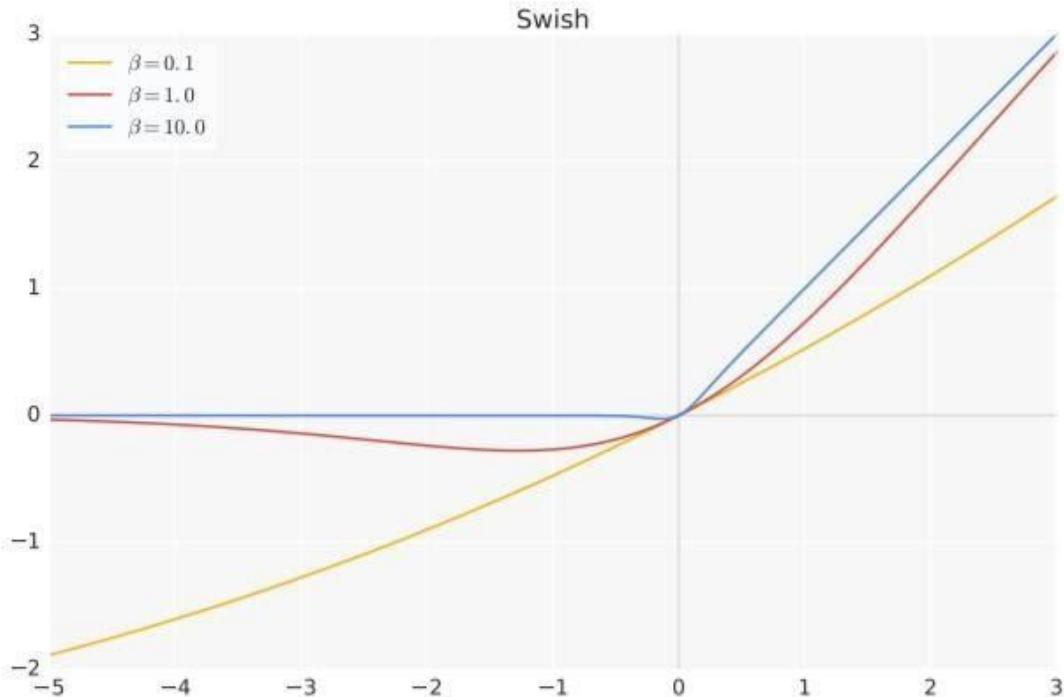

*Fig. 11. Different values of β in Swish function*

A linear function will be attained if β is moved closer to zero; however, the function will look like ReLU if β gets closer to one. However, the beta used in this simulation is among the learned as well as a parameter.

The Swish function is also easily derivable [Eq.19.].

$$f'(x) = f(x) + \sigma(x) - \sigma(x) * f(x) = \frac{e^{-x}(x+1)+1}{(1+e^{-x})^2} \quad (19)$$

This function can deal with vanishing gradient problem that sigmoid is unable to. Moreover, experiments prove that swish is doing better than the ReLU – so far the most efficient activation function of deep learning. It is a fact, the computation of the function has much higher cost than ReLU or even its siblings for both feed-forwarding and backpropagation[44].

## Comparisons and Results Analysis

This section will compare various functions and properties of the activation functions. It will involve numerous properties of the AFs discussed above summary, which will lead to an appropriate conclusion on the paper. Also, real-world applications of the AFs will be considered here other than theoretical analysis. The criteria for this analysis will be based on the speed of training and accuracy of the classification. The table below shows the analysis of the various AFs with their equations and the range. All hidden nodes have the same activation function for each series of tests of given formula, with the exception of the final layer which always uses softmax AF. The network has been implemented using keras and tensorflow frameworks for Python 3. All operations were done on a single GPU unit: Nvidia GeForce GTX 860M.

The experiment was performed with data-set CIFAR-10[45]. The CIFAR-10 data-set consists of 60000 32x32 colour images splitted into ten classes, with 6000 images per class. There are 50000 training images and 10000 test images.Training consists of 25 epochs. The final accuracy is not perfect, nonetheless aim is to compare efficiency of different types of activation functions in the same model, therefore for the experiment a simple network [46] with just two convolution layers has been implemented. In order to alleviate randomness there are three series of each AF application and the final result is the average value for given AF.

*Table 5. Final ratio of correctly classified images*

|  | sigmoid | tanh | relu | leaky_relu | swish | softsign | softplus |
|---|---|---|---|---|---|---|---|
| Final accuracy | 0.6166 | 0.6757 | 0.7179 | 0.7295 | 0.6989 | 0.6901 | 0.6598 |

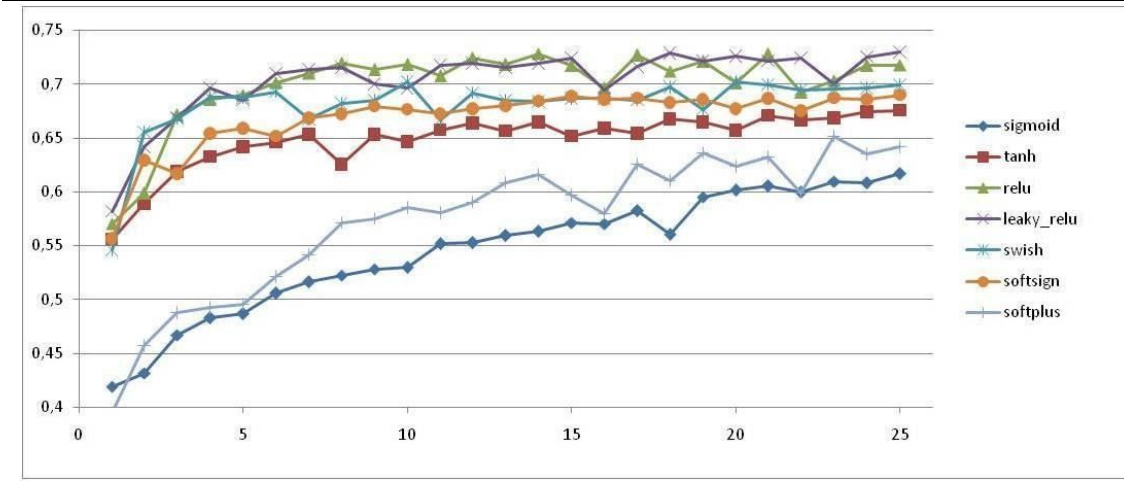

*Fig. 12. Network accuracy over training for each activation function*

Results are shown in 3 diagrams. First one [Fig. 12.] presents the most important parameter: rate how many images were correctly classified. It appears that ReLU and Leaky ReLU were the most successful, since all other networks completed task with less than 70% accuracy [Tab. 5.]. This proves that that ReLU is still reliable and even "dying ReLU" is not able to decline overall performance.

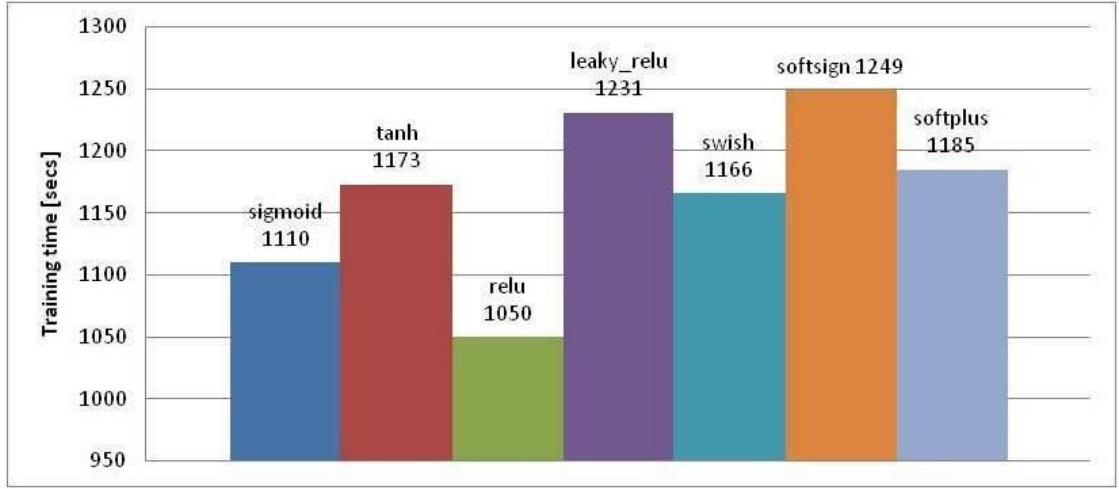

*Fig. 13. Average training time for AF model in seconds*

Regarding time performance, ReLU is unquestionably leader. It took over 5% less time to train than the second one - sigmoid and 15% less time than its sibling - Leaky ReLU [Fig. 13.] . In addition for ReLU-network it took only 2.4 seconds to classify ten thousand images, while the second in row - Softplus almost 2.5 seconds spent on this task [Fig. 14.] . This empirically proves superiority of basic ReLU activation function.

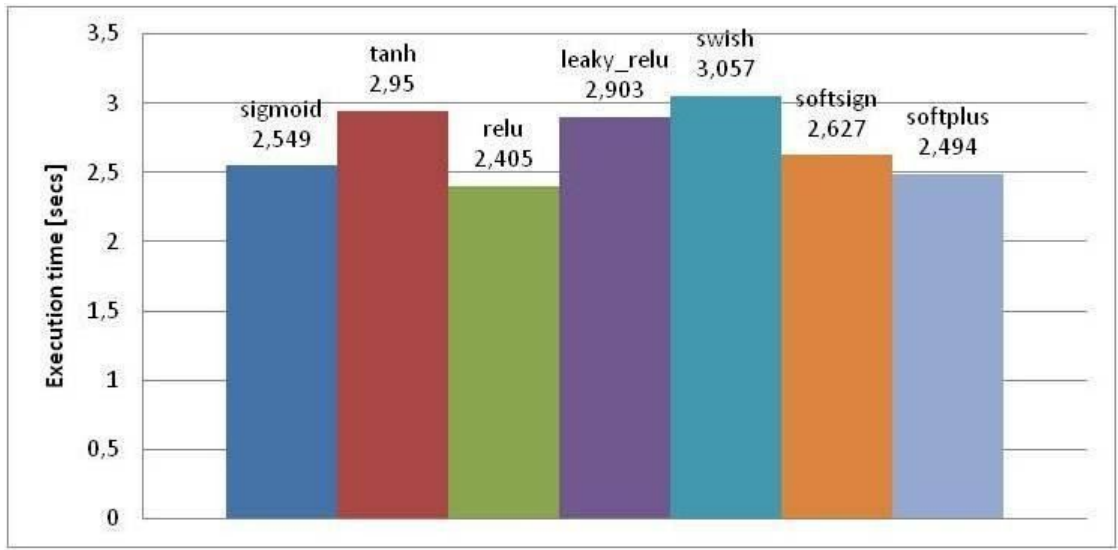

*Fig. 14. Average time to classify 10 000 images with each AF*

# Conclusions

*Table 6. Recommendation when to use which activation function in deep neural networks*

| Function | Comment | When to use? |
|---|---|---|
| Step | Does not work with backpropagation algorithm | Rather never |
| Sigmoid | Prone to the vanishing gradient function and zigzagging during training due to not being zero centered | Can t into Boolean gates simulation |
| Tanh | Also prone to vanishing gradient | In recurrent neural network |
| Softsign | | Rather never |
| ReLU | The most popular function for hidden layers. Although, under rare circumstances, prone to the dying ReLU" problem | First to go choice |
| LReLU | Comes with all pros of ReLU, but due to not-zero output will never "die" | Use only if You expect "dying ReLU" problem |
| Maxout | Far more advanced activation function than ReLU, immune to "dying", but much more expensive in case of computation | Use as last resort |
| Softplus | Similar to ReLU, but a bit smoother near 0. Comes with comparable benets as ReLU, but has more complex formula, therefore network will be slower | Rather never |
| Swish | Same as leaky ReLU, but according to the above research it does not outperform ReLU. Might be more useful in networks with dozens of layers | Worth to give a try in very deep networks |
| SoftMax | | For output layer in classification networks |
| OpenMax | | For output layer in classification with open classes possibility |

This paper shows that there is no ultimate answer for questions like "which activation function should I choose?" However, after this comprehensive summary of activation functions used in deep learning, we can make a few but certain recommendations based on the provided theory as shown in the table above. Thus, this paper summarizes comprehensively the activation functions applied in deep learning (DL)[Tab. 6.], highlighting the current trends in the application of these functions, which is most essential and has never been published in any literature. It started by a presentation of a brief introduction on the activation function and deep learning, which was then followed by analysis of different AFs and a discussion of some applications fields of these functions can be used based on systems and architectures in the development of deep neural networks. Moreover, the activation functions can adjust to the process better or worse since they have the ability to improve the learning of certain data patterns. Therefore, the functions develop over the years and still there is a need to do proper research before applying any of these functions in deep learning for each project separately and this paper is only a guidance.